\pdfoutput=1

\documentclass[11pt]{article}

\usepackage[dvipsnames]{xcolor} 
\usepackage[final]{acl}

\usepackage{times}
\usepackage{latexsym}

\usepackage{tikz}
\def\checkmark{\tikz\fill[scale=0.4](0,.35) -- (.25,0) -- (1,.7) -- (.25,.15) -- cycle;} 
\usepackage{amssymb}
\usepackage{pifont}
\newcommand{\xmark}{\ding{55}}%

\usepackage[T1]{fontenc}

\usepackage[utf8]{inputenc}

\usepackage{microtype}

\usepackage{inconsolata}

\usepackage{graphicx}

\usepackage{hyperref}
\title{Lexical Complexity Prediction and Lexical Simplification for Catalan and Spanish: Resource Creation, Quality Assessment, and Ethical Considerations\thanks{This is a considerable modification to a preliminary version archived in  arXiv.}}

\author{
  \textbf{Horacio Saggion\textsuperscript{1}},
  \textbf{Stefan Bott\textsuperscript{1}},
  \textbf{Sandra Szasz\textsuperscript{1}},
  \textbf{Nelson Pérez\textsuperscript{2}},\\
  \textbf{Saúl Calderón\textsuperscript{2}},
  \textbf{Martín Solís\textsuperscript{2}}
\\
  \textsuperscript{1}Universitat Pompeu Fabra (Barcelona, Spain),\\
  \textsuperscript{2}Instituto Tecnológico de Costa Rica (Cartago)
\\
\\
  \small{
    \textbf{Correspondence:} \href{mailto:horacio.saggion@upf.edu}{horacio.saggion@upf.edu}
  }
}

\begin{document}
\maketitle
\begin{abstract}
Automatic lexical simplification is a task to substitute lexical items that may be unfamiliar and difficult to understand with easier and more common words. 
This paper presents the description and analysis of two novel datasets for lexical simplification in Spanish and Catalan. This dataset represents the first of its kind in Catalan and a substantial addition to the sparse data on automatic lexical simplification which is available for Spanish. Specifically, it is the first dataset for Spanish which includes scalar ratings of the understanding difficulty of lexical items. In addition,  we present a detailed analysis aiming at assessing the appropriateness and ethical dimensions of the data for the lexical simplification task. 
\end{abstract}

\section{Introduction}
Various types of readers may have problems with the understanding of written text. These groups include, among others, language learners \cite{rets2021simplify}, children \cite{javourey2022simplification}, people with cognitive disabilities \cite{licardo2021differences}, and people with a generally low level of reading proficiency. On the other hand, some texts are written in a style that makes it hard to understand the content, for example, by being written in a difficult style or by the use of vocabulary that is unknown to the reader. Universal access to information in the form of understandable text is not only a desirable service to citizens, but it is a citizens' right that has started to be recognized by international institutions and national legislation in the last years.\footnote{For example the plain writing act of 2010: \url{https://www.govinfo.gov/app/details/PLAW-111publ274}} Apart from recognized 
rights, there are also very serious general concerns about inclusion, the principled functioning of democracy and democratic institutions, as well as the right of citizens to be protected from political and economic abuse \cite{rennes2022automatic,johannessen2017review}. Democratic processes have serious shortcomings when certain groups are denied informed participation, just because essential information is not available in a form they can understand. 

A common and effective, although costly strategy to remedy this is to adapt these texts by specialized human editors \cite{nomura2010guidelines}. This approach is 
limited by the vast amount of texts which are available today. 
A 
much more economic alternative is to adapt texts automatically with computational algorithms. This Natural Language Processing task is known as \textit{Automatic Text Simplification} (ATS) \cite{Saggion'2017}. ATS may involve several transformations including sentence splitting, grammatical transformation or the exclusion of overly detailed content. \textit{Automatic Lexical Simplification} (LS) \cite{shardlowsurvey2014,paetzold-specia-2017-lexical} is a well-defined sub-task of ATS, which only aims at finding i) words that are complex and 
should be simplified and ii) simpler substitutes for these complex words. These two sub-tasks are referred to as \textit{Complex Word Identification} (CWI) \cite{zampieri2017complex} and \textit{Substitute Generation} (SG). Finally, \textit{Substitute Ranking} (SR) and \textit{Substitute Selection} (SS) ensure that the best candidate(s) produced by SG are selected for the output. A similar 
task to CWI is \textit{Lexical Complexity Prediction} (LCP) \cite{shardlow2021semeval}, which outputs an estimate for the lexical difficulty of each target unit, instead of only making a binary decision on whether a word should be substituted or not. 

The availability of data that represent LCP and LS is a prerequisite for the development or fine-tuning of models to effectively handle these tasks. Data is needed 
to evaluate and benchmark them. As in the case of many other NLP tasks most work has been done for English. For Spanish the availability of suitable data is low and in the case of Catalan, it is, to the best of our knowledge, nonexistent. The work we present here aims to remedy this situation.

The main contributions of this paper are:

\begin{itemize}
    \vspace{-4pt}
    \item We provide a detailed description of two datasets for \textit{Lexical Simplification} and \textit{Lexical Complexity Prediction} for Spanish and Catalan. 
    \vspace{-6pt}
    \item We describe in full the data compilation process and provide a statistical description of the datasets.
    \vspace{-6pt}
    \item  We assess the quality of the dataset for the lexical simplification task and consider ethical implications of the data.
    \vspace{-4pt}
\end{itemize}

This paper is organized as follows: Section \ref{related} overviews of the state of the art in LS and describes 
existing comparable resources for Iberian Romance languages; Section \ref{sec:method} details the method for data collection and annotation; Section \ref{sec:QA} describes the quality analysis of the data. In Section \ref{sec:ethics} we raise ethical concerns in LS while in Section \ref{sec:conclusion} we close the paper with a discussion and future work.



\section{Related Work}
\label{related}

Foundational work on Lexical Simplification was developed for 
English 
by \citet{Dev:Tai:98} who used Wordnet to identify synonyms for target words and word frequencies from the Kucera-Francis psycho-linguistic database for synonyms ranking. This initial approach was followed by corpus-based approaches that used Language Models \cite{Bel:Moe:10} or Wikipedia \cite{Biran:11,Yatskar10,horn-etal-2014-learning}. Deep learning approaches were explored by \citet{glavas-stajner-2015} with an unsupervised approach for LS based on current distributional lexical semantics modelling, while \citet{paetzold-specia-2017-lexical} combine learned substitutions from a corpus using neural networks. \citet{qiang2020lsbert} presented LS-BERT, a LS framework that uses a pre-trained representation of BERT \cite{devlin-etal-2019-bert} for English to propose substitution candidates with high grammatical and semantic similarity to a complex word in a sentence.

Regarding 
LS in Spanish, few approaches are reported in the literature. They can be classified as: (i) knowledge-based approaches which rely on ``curated'' lists of synonyms and corpora to propose and rank synonyms by relying on frequency and other word characteristics \cite{BottRDS12,Baeza-YatesRD15,ferresSG_BGNLP2017}; (ii) translation-based approaches which cast simplification as translation (\citet{stajner2014} and \citet{STAJNER201980} implicitly learn simplification rules) and (iii) current transformer-based approaches \cite{alarconCTTS2021} which achieve a state of the art performance. In the context of the TSAR 2022 Lexical Simplification challenge \cite{Saggion&al'2022}, several approaches have been proposed, mostly based on pre-trained language models. Controllable lexical simplification was introduced for English in \citet{Sheang&al'2022} achieving state of the art in multilingual settings in \citet{Sheang&Saggion'2023}. Contrary to current methods, \citet{Stajner&al'2023} presents a light-weight text simplifier for Spanish claiming that it achieves good performance without the cost associated with current architectures.  

In the earlier approaches to \textit{Lexical Simplification}, 
CWI was treated as an implicit part of the simplification pipeline, even though it was often 
treated as a modular pipeline component 
\cite{carroll1998practical,shardlow2014out,bott2012can}. \citet{shardlow2013comparison} is the first work which frames CWI as an independent task ``which may seem intuitively easy, but in reality is quite difficult and rarely performed''. He presents a dedicated CWI classifier using Support Vector Machines. 
In 2016 and 2017 two shared tasks were held at \mbox{SemEval} and BEA \cite{paetzold2016semeval,yimam2018report} on CWI. The 2017 task also included an estimation of the probability of a target word being 
complex, which was a step towards \textit{Lexical Complexity Prediction}, but it did not require a direct estimation of \textit{Lexical Complexity}.  ALexS \cite{ortiz2020overview} was a CWI competition for Spanish which unfortunately seldom attracted participants. In 2021, a SemEval shared task invited contributions for LCP \cite{shardlow2021semeval}, which now predicted grades of 
LC directly. This last task was based on previous work in \citet{shardlow2020complex}. The 2024 Multilingual Lexical Simplification Pipeline shared task \cite{Shardlow&al'2024} is a new challenge covering aspects of LCP and LS.

CWI and LCP has been tackled with the use of SVMs \cite{shardlow2013comparison}, decision trees \cite{quijada2016hmc}, random forests \cite{ronzano2016taln} and neural networks \cite{nat2016sensible}. Recent approaches include the use of transformer models \cite{yaseen2021just}.  

As for the coverage of Spanish and Catalan \citet{ferres2017adaptable} presents a CNN classifier for CWI in Portuguese, Spanish, Catalan and Galician and \citet{sheang2019multilingual} builts a multilingual system based on a CNN and linguistic feature engineering for multilingual CWI, which covers Spanish, English and German. So far, these systems tackled CWI, but not LCP with predictions on a complexity scale.

 \label{rescources_iberian}
 
Concerning LS datasets, the aforementioned shared tasks produced valuable resources, mainly for English. There exist LS datasets for Portuguese \cite{hartmann2018simplex} and Japanese \cite{kodaira2016controlled}. \citet{uchida2018cefr} present a dataset for the the educational domain. 

\begin{table*}[h!]
\scriptsize
 \begin{center}
\begin{tabular}{c|c|c|c|c|c|c}
      
      \multicolumn{7}{c}{Catalan} \\
      \hline
       
       & Av & Av Years & Av Reading  & \#Partici- & \#Native & Languages\\
     Annotators        & Age & in Education & Hrs per Week &    pants    &  Speakers & Spoken (L2)\\ 
         
  \hline
        Personal &    58.21 (14.36)  & 17.93 (4.89) & 10.21 (10.54)  & 14  &  8 & 2.21 (1.25)\\ 
      \hline
        Prolific &  29.30 (8.54) & 16.98 (3.24) &  7.17 (6.06)   &  60     &  13  & 2.08 (0.81)\\
      \hline
        All      &  34.77 (15.02)  &  17.16 (3.59)  &  7.75 (7.14) & 74  &  21  & 2.18 (0.90) \\  
        \hline
          \multicolumn{7}{c}{} \\
        
      \multicolumn{7}{c}{Spanish} \\
      \hline
         & Av & Av Years & Av Reading  & \#Partici- & \#Native & Languages\\
     Annotators        & Age & in Education & Hrs per Week &    pants    &  Speakers & Spoken (L2)\\ 
     \hline
        
        Personal &  34.50 (13.42)  & 21.78 (3.31)  &  14.00 (17.35)  &  10  & 7 &  4.1 (2.00) \\
      \hline
        University &   17.98 (1.38) & 12.16 (1.50)  & 2.73 (2.80) & 60 & 60 & 1.93 (0.55) \\
      \hline
        All &   22.11 (10.85)  & 13.69 (4.21) & 5.67 (14.59) & 70 & 67  & 2.31 (1.05) \\ 
      \hline
\end{tabular}
\end{center}\caption{Demographic statistics on participants in the data collection. Standard Deviation is given in parentheses. \textit{Personal} stands for personal contacts, \textit{university} for university students and \textit{prolific} for platform annotators.}
\label{democraphCat}

\end{table*}

For Iberian Romance Languages, to the best of our knowledge, there are only two datasets for LS in Spanish: EASIER and ALEXSIS.
The EASIER dataset was used for CWI and SG/SS tasks \cite{alarconIEEEAccess2021}; it contains about 5,130 instances \cite{alarconCTTS2021} with at least one proposed substitute per complex word. A smaller portion of the dataset which contains 575 instances is more realistic for LS since it contains three proposed substitutes, although without ranking.  The EASIER-500 dataset containing 500 instances\footnote{\url{https://data.mendeley.com/datasets/ywhmbnzvmx/2}}
was used to evaluate SG and SS approaches \cite{alarconIEEEAccess2021,alarconCTTS2021}.  ALEXSIS \cite{alexsisLREC2022} contains 381 instances composed of a sentence, a target complex word, and 25 candidate substitutions. For every pair $<$sentence, complex word$>$ a simpler substitute was annotated by a set of 25 annotators. The sentences and complex words of this dataset were extracted from the CWI Shared Task 2018 dataset\footnote{\url{https://sites.google.com/view/cwisharedtask2018/datasets}} for Spanish \cite{yimametal2018} being its format similar to that of LexMturk \cite{horn-etal-2014-learning} for English.
Again, these datasets cover CWI, but not LCP. In the case of Catalan, there are, to the best of our knowledge, no available datasets at all.

\section{Methodology of the Dataset Creation}
\label{sec:method}

Both datasets have been created within the data collection efforts for a lexical simplification  
shared task  \cite{Shardlow&al'2024}. The target selection and data collection process of the datasets for Spanish and Catalan was largely parallel, but there were some differences due to the availability of source texts and annotators. The initial goal was to select 600 target words per language in 200 contexts, with 3 targets per context. An additional 10 contexts (and 30 words) were required for pilot annotations. Due to the sparseness of resources we had to relax the goal for Catalan to 160 contexts. For each target a minimum of 10 annotations was required which were collected through on-line forms.

The annotation process collected two pieces of data for each target word: i) a rating on Lexical Complexity on a 5-point Likert scale 
(from "very easy" to "very hard")
and ii) up to 3 lexical substitutes for the target that fit in the given context. Annotators were asked to simply repeat the target word if they could not find a suitable alternative.

In addition to the annotation itself, participants were asked to give some 
demographic data for the creation of simple statistics:  age, years in education,  average hours per week used for reading,  whether the participant was a native speaker, 
the number of languages spoken and their native language. Education and weekly reading can be seen as proxies for stylistic and language proficiency and may be used in future studies.
Personal data was stored anonymously and separate from annotation data and any data which would allow inferences on the identity of participants was deleted after the dataset compilation. Table \ref{democraphCat} gives the resumed demographic information about the participants.

The structure of the datasets is similar to the one of ALEXSIS (described in Section \ref{rescources_iberian}), with two important differences: (1) ALEXSIS only contains words for which at least one lexical simplification could be found by the annotators, (2) target words in ALEXSIS do not contain lexical complexity values. Concerning the first point, our datasets also provide examples of non-substitutable words, which is also important for system developments. 

The datasets presented here correspond to a combined scenario. This will help the development and assessment of systems that jointly or separately address the lexical simplification pipeline \cite{paetzold-specia-2017-lexical}. 
The average ratings on Lexical Complexity are listed normalized to a scale from 0 to 1. 
Repeatedly proposed substitutions are listed as many times as they were proposed by different annotators. This implies a non-monotonic ranking of their preference. 
An example of a Catalan and a Spanish annotation is shown in Table \ref{DSexamples}.

\begin{table*}[h!]
\scriptsize
 \begin{center}
\begin{tabular}{p{0.7in}|p{5in}}
      \hline
          \\
 	 \small{Spanish Example} &    \multicolumn{1}{|p{4in}}{{\em \footnotesize{
 	 Pero uno no puede dejar que el {\bf derrotismo}  {\em \footnotesize{\em lo detenga e impida que haga un presupuesto}}}}}\\ \\
	\small{LC of target}	  &\small{0.7} \\ \\
	\small{Substitutes}	  &\footnotesize{desánimo (4), pesimismo (4), abatimiento (3), derrotismo (2), desesperanza (1), desaliento  (1), catastrofismo (1), negativismo (1)} \\ \\
  	  \hline
      \hline
    \\
 	 \small{Catalan Example} &    \multicolumn{1}{|p{4in}}  {\em \footnotesize{
 	 No poden tocar-se ni abraçar-se, no hi ha joc col·lectiu, s'ha  {\bf sectoritzat}  \em{\em el pati i la desinfecció per allà on passen és la nova rutina a l'escola.}}}\\ \\
	\small{LC of target}	  &\small{0.6} \\ \\
	\small{Substitutes}	  &\footnotesize{dividit (5),	segmentat (2), fragmentat (1), seccionar (1), sectorizat (1), divisió en sectors (1), sectoritzat (1), senyalitzat (1), compartimentat (1), dividit en parts (1), en grups (1), classificat (1), separat en zones (1)} \\ \\
    \hline
\end{tabular}
\end{center}
\caption{Examples from our datasets with complexity ratings and LS substitutes. The count of how many times the same word was proposed by different annotators is given in parentheses here, while in the datasets it is represented by the repetition of the words.}
\label{DSexamples}
\end{table*}

\subsection{Catalan Dataset}\label{catDS}

The Catalan dataset consists of 160 context sentences containing 475 target word tokens (454 distinct types). Sentences were selected from the Educational news section of the TeCla corpus\footnote{\url{https://huggingface.co/datasets/projecte-aina/tecla}} \cite{armengol-estape-etal-2021-multilingual} of 
news texts. 

\subsubsection{Data Preparation}

A first pre-selection of candidate \textit{contexts} was done with an automatic process that selected all sentences containing a minimum of 3 content words above a frequency threshold on lemma counts. This threshold was used as an approximate criterion of word difficulty. The frequency was measured with the Catalan Spacy\footnote{\url{https://spacy.io/models/ca}} model. 
The selected contexts were then randomized in order and presented to two annotators (proficient L2 speakers) who had to decide for each word if it was a good simplification candidate because it i) was a complex word and ii) potentially any substitutes could be found for it. 

\subsubsection{Data Selection: Target Words and Context Sentences}

Based on this pre-annotation, we selected target contexts that contained at least one target word unit on which both annotators agreed. For each context 3 targets were selected, giving first preference to units that were agreed on as being complex by the annotators, then those which were marked by only one of them. We did this in order to include words which are guaranteed to be complex and simplifiable. As a last resort, an infrequent word could be selected at random if less than 3 manually marked complex words were available in a sentence. This also allowed the inclusion of some words which might potentially not be simplifiable. This process gave us a total of 480 target words, embedded in 160 context sentences, with each context containing 3 targets. This data was divided into batches (3 batches of 10 targets for a pilot annotation 
and 9 batches of 50 targets for the rest). Each batch was annotated by a fixed set of annotators.

\subsubsection{Annotation}

Target words were annotated by proficient Catalan speakers (see Appendix A) 
We monitored the annotation process in Prolific to detect workers not following the annotation guidelines. For example, annotators who always returned 
target words as substitutes or provided synonyms in Spanish were contacted and allowed to re-annotate if they wanted. 
Finally, we had to reject 11 annotators. Of the target words 5 had to be removed because they were not correctly presented to the annotators or did not potentially have a meaningful substitute (e.g. calendar dates).

\begin{table*}[ht]
\scriptsize
    \centering
    \begin{tabular}{l|c|c|c|c|c|c|c|c|c} 
    \multicolumn{10}{c}{Spanish} \\ \hline
        LC Level & \multicolumn{2}{|c|}{validity} &  \multicolumn{2}{|c|}{equivalence} &  \multicolumn{2}{|c|}{in-context fit} &  \multicolumn{3}{|c}{simplicity} \\ 
        & V & NV & E &  NE & F & NF & S & EQ & C \\ \hline
        1 $[0.00..0.20]$ & 100\% & 0\% & 87\% & 13\% & 100\% & 0\% & 35\% & 50\% & 15\% \\        \hline
        2  $(0.20..0.40]$ & 100\% & 0\% & 87\% & 13\% & 81\% & 19\% & 42\% & 50\% & 8\% \\        \hline
        3  $(0.40..0.60]$ & 100\% & 0\% & 63\% & 37\% & 79\% & 21\% & 42\% & 58\% & 0\% \\        \hline
          4  $(0.60..0.80]$ & 100\% & 0\% & 77\% & 23\% & 74\% & 26\% & 65\% & 35\% & 0\% \\        \hline
           5  $(0.80..1.00]$ & 100\% & 0\% & 73\% & 27\% & 86\% & 14\% & 59\% & 41\% & 0\% \\        \hline
          ALL & 100\% & 0\% & 77\% & 23\% & 84\% & 16\% & 48\% & 46\% & 6\% \\        \hline
    \end{tabular}

 \begin{tabular}{l|c|c|c|c|c|c|c|c|c} 
   \multicolumn{10}{c}{Catalan} \\ \hline
       LC Level & \multicolumn{2}{|c|}{validity} &  \multicolumn{2}{|c|}{equivalence} &  \multicolumn{2}{|c|}{in-context fit} &  \multicolumn{3}{|c}{simplicity} \\ 
        & V & NV & E &  NE & F & NF & S & EQ & C \\ \hline

            1 $[0.00..0.20]$ & 100\% & 0\% & 77\% & 23\% & 74\% & 26\% & 26\% & 61\% & 13\% \\        \hline
          2   $(0.20..0.40]$ & 97\% & 3\% & 93\% & 7\% & 70\% & 30\% & 44\% & 56\% & 0\% \\        \hline
         3  $(0.40..0.60]$ & 100\% & 0\% & 70\% & 30\% & 76\% & 24\% & 62\% & 38\% & 0\% \\        \hline
           4 $(0.60..0.80]$ & 93\% & 7\% & 71\% & 29\% & 75\% & 25\% & 45\% & 45\% & 10\% \\        \hline
          5 $(0.80..1.00]$ & 100\% & 0\% & 67\% & 33\% & 100\% & 0\% & 50\% & 50\% & 0\% \\        \hline
        
         ALL &  97\% & 3\% &  78\% & 22\% & 58\% & 42\% &  44\% & 50\% & 6\% \\        \hline
 
    \end{tabular}

    \caption{Qualitative Assessment of the Analysed Substitutes in Spanish and Catalan by complexity level and overall. V: valid word, NV: not valid word, E: equivalent word, NE: non equivalent word, F: fit in context, NF: not fit in context, S: simpler, EQ: equaly simple/complex, C: more complex.} 
    \label{tab:QUAN_results_COMBINED}
\end{table*}

\subsection{Spanish Dataset}\label{spaDS}

The Spanish dataset consists of 625 target words in 210 contexts from texts on educational books on finance (see also Appendix B). 

\subsubsection{Data preparation}

Our lexical simplification dataset for Spanish derives from a corpus of over 5K sentences for sentence simplification currently  under development. The sentences  were simplified following a set of simplification guidelines borrowed from the Simplext project \cite{saggion2015}.
Each sentence was simplified by one of six annotators who were trained to follow the simplification guidelines. 
The corpus features interesting simplification phenomena such as the transformation of numerical  information ({\it 10\%} $\rightarrow$ {\it diez por ciento})  -- a well known simplification operation \cite{Bautista&Saggion'2014}, the splitting of a long sentence into two shorter ones, and lexical substitutions ({\it derrotismo 
} $\rightarrow$ {\it pesimismo}). 

\subsubsection{Data Selection: Target Words and Context Sentences}

Lexical simplification candidates were heuristically mined from the corpus in order to create our novel 
LS dataset for Spanish. We search specifically for sentence pairs in which a word was present in the original complex sentence but missing in the simplification. 
A Natural Language Processing pipeline for Spanish\footnote{\url{https://spacy.io/models/es}} was used to analyze original and simplified sentences and extract words and parts-of-speech tags. We restricted our analysis of lexical simplification to single content words with POS tags noun, verb, adjective or adverb, excluding Multi Word Expressions. The set of unique words in the original and simplification was compared to assess whether a {\it complex} $\rightarrow$ {\it simple} transformation could be identified.  A transformation {\it complex} $\rightarrow$ {\it simple} was considered a priory valid substitution if the pair of words were semantically related and not a morphological derivation of one another.  A semantic similarity threshold and a lexical similarity threshold were computed in order to implement this validation check using the test data from the ALEXSIS dataset to adjust parameters (see Section~\ref{rescources_iberian}): all pairs of complex words and substitution words in ALEXSIS were compared using cosine similarity in a Spanish Word Embedding space\footnote{Large Spanish Fasttext Word Embedding model \url{https://zenodo.org/records/3255001}} and the cosine values averaged to obtain a similarity threshold (i.e. similarities greater that the threshold used as an indication of word relatedness).  A second value was computed to discard morphological similar (e.g. {\it obtenido} and {\it obtener}) pairs: the edit distance between candidates was computed and averaged over all ALEXSIS pairs.  These two thresholds were used as a means to discard complex sentences containing a word without an equivalent simplification in the simple sentence, for example, in cases where the sentence underwent a delete operation or 
a different verb form was used in the simplification. 
With this, 
we obtained 1,533 complex sentences 
containing a potential target word, that is a word which was replaced by a related word in the simplification. This set provided the basis for the human annotation of the dataset.

The selected words in their sentence context were annotated by two annotators (one native Spanish speaker and one with Spanish as L2) on whether the word in question was a good simplification target (being complex and potentially "simplifiable"). In case of doubt dictinonaries were consulted. 
The process yield 601 valid contexts -- contexts were at least one target word on which both annotators had agreed. The data was analyzed again to extract two additional content words from each sentence to provide words which 
could potentially be 
"non-simplifiable". From this set, we sampled 210 target contexts by taking into account the average sentence length, selecting sentences whose length deviated at most one standard deviation from the mean length.   
We ensured 
that each target word only appeared once in the dataset as a target. 

\begin{table*}[h!]
\scriptsize
 \begin{center}
\begin{tabular}{l|p{5in}} \hline

Target / Substitute & Sentence with target / Sentence with substitute / Sentence with correct  \\ \hline 

 Tgt: {\bf mercancías} (LC: 0.3) & \checkmark \hspace{0.20cm}  El mercado es el lugar donde se transan las {\bf mercancías} y los servicios; es la expresión que define el lugar físico o figurado donde se encuentran vendedores y compradores. ({\em The market is the place where {\bf goods} and services are traded; It is the expression that defines the physical or figurative place where sellers and buyers meet.})\\  \\

Sbs: {\bf productos} &  \xmark \hspace{0.2cm} El mercado es el lugar donde se transan \textcolor{red}{las} {\bf productos} y los servicios; es la expresión que define el lugar físico o figurado donde se encuentran vendedores y compradores. \\ \\

 & \checkmark \hspace{0.2cm} El mercado es el lugar donde se transan \textcolor{blue}{los} {\bf productos} y los servicios; es la expresión que define el lugar físico o figurado donde se encuentran vendedores y compradores. \\ \\

\hline

\end{tabular}
\end{center}
\caption{Substitution Amendment Examples in Spanish. In \textcolor{red}{red} we highlight the problems when the substitute is used as a direct replacement and in \textcolor{blue}{blue}  how it can be amended. Target = Tgt, Substitute = Subs.} 
\label{tab:Subs_Context_ES}

\end{table*}  

\begin{table*}[h!]
{\scriptsize
 \begin{center}
\begin{tabular}{l|p{5in}} \hline
Target / Substitute & Sentence with target / Sentence with substitute / Sentence with correct  \\ \hline 
Tgt: {\bf manifest} (LC: 0.39) & 
\checkmark \hspace{0.20cm}  ... es va crear una comissió de seguiment que s'ha anat reunint d'aleshores ençà i a l'entorn de la qual es van posar de {\bf manifest} algunes mancances ... ({\em ... a follow-up commission was created which has been meeting ever since and around which some shortcomings became {\bf manifest} ...})  \\ \\

Sbs: {\bf evidència} & \xmark \hspace{0.20cm} ... es va crear una comissió de seguiment que s'ha anat reunint d'aleshores ençà i a l'entorn de la qual es van posar \textcolor{red}{de} {\bf evidència} algunes mancances ... \\  \\
      
& \checkmark \hspace{0.20cm} ... es va crear una comissió de seguiment que s'ha anat reunint d'aleshores ençà i a l'entorn de la qual es van posar  \textcolor{blue}{en} {\bf evidència} algunes mancances ... \\  \\ \hline

\end{tabular}
\end{center}
\caption{Substitution Amendment Examples in Catalan. In \textcolor{red}{red} we highlight the problems when the substitute is used as a direct replacement and in \textcolor{blue}{blue}  how it can be amended. Target = Tgt, Substitute = Subs.}
\label{tab:Subs_Context_CAT}
}
\end{table*}  

\subsubsection{Annotation}

The resulting 630 target words were divided into a first batch of 30 contexts and target words to run a trial annotation and a batch of 200 contexts and target words to produce the final dataset. This task was undertaken by students who are native Spanish speakers and by social contacts of the authors. The trial annotation was done by personal contacts, while the main part of the dataset was annotated as part of a curricular activity. 

Each data point was annotated by 10 participants. Five data points had to be removed, 3 of them because no meaningful synonyms could be found (e.g. URLs) and two because there was an error in the annotation forms which prevented participants from giving meaningful answers. So the final dataset consists of 625 target words in 210 contexts. 

\subsection{Lexical Complexity Analysis}

Lexical Complexity is perceived quite subjectively, although some factors, e.g. word frequency in day-to-day communication, are relatively objective factors, despite the fact that corpora may not always represent the day-to-day exposure of language to individuals faithfully. So, one important and interesting question is in how far different annotators agree in their complexity judgements. We expected to find a relatively strong, but not perfect agreement among raters. To assess inter-annotator agreement on complexity rating, it has to be considered that the values from the Likert-scale are ordinal and fall on an interval scale. The best way to treat this is by calculating agreement on the ranking of rated items. For this reason, we use Intraclass Correlation Coefficient (ICC) and Spearman's rho.
ICC estimates were calculated using Pingouin \cite{vallat2018pingouin} statistical package version 0.5.4 based on a mean-rating, one-way random effects multiple raters model (ICC1k) \cite{shrout1979intraclass}. ICC values were calculated for each annotation batch (for which the set of raters was fixed) and then averaged. ICC1k was 0.78 for Spanish and 0.62 for Catalan. There is no generally accepted way to interpret ICC scores, but the value for Spanish can be described as \textit{good} and the one for Catalan as \textit{moderate} \cite{koo2016guideline}. In the light of what we said above, this is an expected result.

\section{Dataset Quality Analysis}
\label{sec:QA}

In order to assess the quality of the datasets, we examined several contexts, target words and substitutes to check if those substitutes were  simpler, meaning preserving, and fit for the context when used to replace the target word in the given context. While doing our analysis, we  considered the top three (most frequent) suggested substitutes per target word hypothesising that they would satisfy the annotation requirements (see Section \ref{sec:method}).  We discover that, although a majority satisfy the desired properties, there is a considerable number of cases which do not comply with  being appropriate in-context substitutes.   

Our analysis consists on examining a sample of 270 data-points: 150 data-points for Spanish and 120 data-points for Catalan. The analysis is carried out by two native speakers of Spanish who additionally have C1 and B2 Catalan proficiency. For the assessment of the data-points speaker linguistic proficiency and knowledge of the language was considered while checking on dictionaries \footnote{For Spanish the Dictionary of the Spanish Royal Academy and for Catalan the Optimot and Diec2 dictionaries.} to reinforce decisions. The method used for selecting the candidates was as follows: First the lexical complexity (LC) level of target words in the datasets was used to create five categories for analysis as shown for example  in Table \ref{tab:QUAN_results_COMBINED}. From each category we selected 10 sentences and their targets (times   their three top most human proposed substitutes). All categories in the Spanish dataset have at least ten sentences to select from by random sampling. As for Catalan, all categories, except category number 5, had enough sentences to sample from randomly. For category 5, we just selected the only sentence in that category. 

The variables of interest for the analysis are as follows: (1) {\em validity} - whether or not the substitute is a valid word in the language (e.g. occurs in a dictionary or is a valid  morphological derivation of a valid word); (2) {\em equivalence} - whether the substitute is equivalent to the target word; (3) {\em in-context fit}: whether the substitute can be used in the syntactic context as the target word; and (4) {\em simplicity} - whether the substitute is less complex, equally complex, or more complex word.  Table \ref{tab:QUAN_results_COMBINED}  presents the overall quantitative results of the analysis as well as the results per lexical complexity category. By looking at the tables we can observe  for the Spanish dataset  that all proposed substitutes analysed are valid words of the language, however just 77\% were considered as equivalent to the target word. Of those considered equivalent an overwhelming  majority (84\%) were considered to directly fit in the context while about half (48\%) were assessed as simpler and 46\% considered as equally complex (or simple).  A trend can be perceived when looking at the analysis per lexical complexity categories, as complexity of the targets increases, the substitutes' equivalence and context fit decrease. A different trend can be observed with respect to simplicity, as the complexity of the target increases also does simplicity of the substitute.  Contrary to the Spanish case, not all Catalan substitutes were valid words in the language (97\% are valid words), however an overwhelming majority (78\%) are equivalent to the target word but with only 58\% being fit for direct replacement. As for simplicity, only 44\% are considered simpler than the target. Looking at the complexity levels,  the picture is not as clear for Catalan, and we speculate that differences with Spanish  may be attributed to the target population who provided the crowd sourced substitutes (i.e. main language Spanish and knowledge of Catalan as second language, see Table \ref{democraphCat}). 

We provide several examples of our analysis that qualitatively illustrate issues related to substitutes which are semantically unrelated, incorrect, or too specific to be used as replacements.

For example, in context "cifras {\bf millionarias} de dinero" ({\em {\bf millionaire} figures of money}) the substitute "acaudaladas" (wealthy) was considered not equivalent since it is an adjective which is used to qualify people and not to qualify abstract concepts such as "cifras" (figures of money). 

Another example would be "...salario o sueldo que se percibe, cuando se tiene un empleo, {\bf honorarios}  que  se cobran como prestaciones de servicios..."  ({\em ... salary that is received, when one has a job, {\bf honoraries} that are charged as services...}), in this case the proposed substitute "pagos" (payment) was considered  non equivalent to "honorarios" (honoraries), the reason being an error in the gender of the word: although "pago" (payment) is a valid word,  in Spanish it is the feminine  "pagas" (wages) which could have been accepted as replacement.
Finally, in the context "hay indicadores financieros que entregan información sobre el pulso {\bf bursátil}, el número y los montos de las transacciones de acciones de sociedades" ({\em There are financial indicators that provide information about the pulse of the {\bf stock market}, the number and amounts of transactions in company shares.}) the proposed substitute "bancario" (banking) is a term referring to the banking domain, too specific to be considered equivalent to "bursàtil" (stock market) which is a broader term (which includes the banking domain).

As for Catalan, we illustrate three examples of incorrect substitutes due to problems of figurative language use or domain connected or semantically related -- but not equivalent -- words.

In the following context:
"El Síndic també posa de manifest que una {\bf sobreoferta} té efectes negatius sobre la segregació escolar..." ({\em The Ombudsman also points out that an {\bf oversupply} has negative effects on school segregation...}) a substitute "sobresaturació" (oversaturation) would not provide a valid replacement for "sobreoferta" (oversupply) since this candidate substitution refers (figuratively) to people undergoing stress and it does not refer to an increase in (educational)  course offer.  

The context " l’Associació Celíacs de Catalunya ha denunciat la “situació d’indefensió” en la que es troben els 30.000 alumnes celíacs o sensibles al {\bf gluten} que mengen als menjadors catalans" ({\em the Associació Celíacs de Catalunya has denounced the "helpless situation" in which the 30,000 celiac or {\bf gluten}-sensitive students who eat in Catalan canteens find themselves})
 the candidate "ségol" (rye) can not be considered a valid replacement since "gluten" (gluten) is a proteine found in cereals like rye, but the terms are not equivalent.

 Finally, in the context "Mitjançant la psicologia, el '{\bf mindfulness}' i el ioga, els alumnes aprenen a resoldre conflictes i, alhora, valors com l'autoestima o el respecte." ({\em  Through psychology, {\bf mindfulness} and yoga, students learn to resolve conflicts and, at the same time, values such as self-esteem or respect. }) the word "meditació" (meditation) cannot be taken as an equivalent of "mindfulness" since these are two different but related concepts in psychology.

In Tables \ref{tab:Subs_Context_ES}  and \ref{tab:Subs_Context_CAT}, we present examples of substitutes which are equivalent to the targets but nonetheless their use as direct replacement is not without consequences for the correctness of the resulting sentence. Indeed, a lexical simplification system should take into account context modification at the local and global level to guarantee grammaticality, coherence, and cohesion. We can observe that gender and governed prepositions have to be adapted to the substitution.





\section{A Note on Ethical Considerations for Lexical Simplification Datasets}
\label{sec:ethics}

Although a very detailed analysis of the dataset could not be carried due to limited resources, we believe it is important to highlight 
aspects related to ethics which have not been addressed thus far in the field of lexical simplification. Since lexical simplification aims at substituting lexical items that may be unfamiliar and difficult to understand,  the automated process may produce output which could raise concerns from the  ethical viewpoint since the replacements may lead to unfair, unethical or false description of people or events. The following is a clear example of discriminatory, offensive language: Let's suppose we are given the sentence "She has a disabled brother." and the target word "disabled".  English dictionaries list "retarded" as an offensive synonym of disabled, therefore in case a system does not take into account that metadata information, an offensive sentence could be produced as in "She has a retarded brother."\footnote{In Spain pejorative terms were recently removed from the Spanish Constitution \href{https://www.lamoncloa.gob.es/serviciosdeprensa/notasprensa/presidencia-justicia-relaciones-cortes/Paginas/2024/180124-congreso-aprobada-reforma-constitucion.aspx}{{https://www.lamoncloa.gob.es/serviciosdeprensa/notasprensa/ presidencia-justicia-relaciones-cortes/Paginas/2024/180124-congreso-aprobada-reforma-constitucion.aspx}}.

} .The same goes without saying for the use of word-embedding models or LLMs which are trained on data which is not properly annotated for ethics.  

The subset of data points we have analyzed already contains some traces of the problems described above, somehow concerning because it directly comes from human informants. Although only a few items entail ethical concerns  a process of carefully revision and ethical disclosure as the one we have put forward here is necessary, specially in the case of a crowd annotated dataset, to understand the risk the provided data  may entail. From a pure automated evaluation  viewpoint, in the previous illustration if the offensive term is used to replace a non-offensive one, being  considered valid in the gold standard, the system producing such output would be rewarded (!).

Although in the Spanish data no serious problems were detected, two relevant cases are present in the Catalan data:  The first case is a replacement suggested by a crowd workers  which  could be considered an euphemism and  which, in this particular case, should be avoided: For the sentence  
"En una segona part, explica Campàs, els participants aprenen estratègies per abordar la violència masclista i que comportaments ``poc visibles'', a la llarga es poden traduir en ``assetjaments, violacions i {\bf feminicidis}''." ({\em In a second part, explains Campàs, the participants learn strategies to address male violence and that ``not very visible'' behaviors, in the long run, can translate into ``harassment, rape and {\bf femicide}''}) and the target word {\bf feminicidi} (femicide), the substiture "assassinat" (murder) was proposed which does not carry the very meaning of the target word also diminishing its intended meaning. 

The second case illustrate the proposal of two offensive terms: For the context 
"Alguns dels alumnes de 5è, amb qui també s’ha treballat una de les cançons del conte encara que no participen a la cantata, han explicat com mai abans havien sentit abans paraules com transsexual o {\bf lesbiana}." ({\em Some of the 5th grade students, with whom one of the songs in the story was also worked on even though they do not participate in the cantata, explained how they had never heard words like transsexual or {\bf lesbian} before.}) and the target word {\bf lesbiana} (lesbian),  one crowd annotator suggested the word "gallimarsot"\footnote{Zoomorphism  to refer to a female who acts as a male. \url{https://dlc.iec.cat/}} while another annotator proposed  the term "marieta"\footnote{Despective for homosexual. Diccionario LGBT+ Catalán  \url{https://lgbt.fandom.com/es/wiki/Diccionario_LGBT}} which can be considered pejorative terms to refer to a lesbian. But note that this example is also interesting in that the term "lesbiana" in this context is referring to the word itself, and therefore should not be replaced.

\section{Conclusions and Future Work}
\label{sec:conclusion}
As we have argued throughout the paper, there is a clear need to have more resources like the one presented here for Catalan and Spanish. Such datasets are a pre\-requisite for the development and evaluation of LS and LCP systems. We have described two novel datasets which  allow the development and evaluation of  Lexical Simplification Systems for Catalan and Spanish. We expect that these datasets are a valuable addition to the currently sparse data in this field. We have quantitatively and qualitatively assessed the dataset confirming the suitability of the dataset for lexical simplification research. Moreover we have also discussed ethical issues discovered through this analysis which should inform further dataset releases. The dataset has already been used in a shared task in lexical simplification \cite{Shardlow&al'2024} and our future work will consider a thorough analysis of system contributions, and in particular how to leverage system outputs to improve data creation and assessment.  Given that target users of text simplification systems include vulnerable populations, we would like to launch {\em a call to arms} for better ethical control during data creation and annotation and evaluation of automatic systems so as to flag at early stages any sensitive issues which may affect the intended user of these systems.

\section*{Lay summary}

For many people accessing information in written texts is too difficult, because the text is written in a style that is too hard for them. This can happen to elderly people, language learners and people with cognitive impairments, among others. Automatic Text Simplification can help to adapt texts for them. Lexical 
Simplification is one aspect of Text Simplification. It replaces difficult words with easier ones. For the creation of Automatic Text Simplification data sets are necessary which contain examples of good substitutions of words with simpler alternatives. We present two datasets of this type for Spanish and Catalan. For Spanish, there are only very few existing datasets so far and for Catalan there are none. Our contribution fills this gap and will make the development of Spanish and Catalan Text Simplification systems possible.

\section*{Acknowledgments}

This document is part of a project that has received funding from the European Union's Horizon Europe research and innovation program under the Grant Agreement No. 101132431 (iDEM Project). Views and opinions expressed are however those of the author(s) only and do not necessarily reflect those of the EU. Neither the EU nor the granting authority can be held responsible for them. We also thank the Departament de Recerca i Universitats de la Generalitat de Catalunya (ajuts SGR-Cat 2021) for its support. 

\bibliography{LS.bib}

\section*{Appendix A: Selection criteria for annotators}

For Catalan, annotators were in part recruited from persons of the social environment of the authors and in part from workers recruited over the Prolific\footnote{\url{https://www.prolific.com/}} crowdsourcing platform.\footnote{Annotators received a fair pay.} All trial data was annotated by social contacts, as well as a part of the main annotations.
In the case of Catalan it is difficult to select a pool of participants that consists only of native speakers because Catalonia is a largely bilingual territory. However, since Catalan has been used as the main vehicular language in the school system for several decades, most people who had their education in Catalonia have a high level of Catalan proficiency. Also a large part of the population grew up bilingually.  

For Spanish, 
the trial annotation was done by personal contacts, while the main part of the dataset was annotated as part of a curricular activity within a course on written communication. This course was designed to foster the development of skills necessary for writing scientific and academic texts that are comprehensible to a broad audience. 
It required the texts to adhere to standards of clarity, precision, coherence, and readability, aligning with the principles of effective scientific communication. The primary intent behind this task was to enhance the student’s ability to identify and modify the use of complex terminology, opting for more accessible alternatives without compromising the accuracy or depth of the content. This approach facilitates widespread dissemination and understanding.

The annotators recruited from personal contacts were mostly speakers of European Spanish    , while the rest were speakers of the Costa Rican variety of Spanish.

Since the availability of annotators was limited, the main criterion for the recruitment of annotators from the personal contacts of the authors was their availability, both for Spanish and for Catalan. We made sure that all of them were proficient speakers of the language, either native or L2 speakers which use the language on a daily basis. Even without having any stricter selection criteria, in practice their annotations were much more reliable than annotations from crowdsourcing workers. For Catalan we had to discard 11 crowdsourcing annotators. 

\section*{Appendix B: Selection criteria for texts}

Both of datasets have been created within the context of the MLSP24 (Multilingual Lexical Simplification Pipeline) shared task \cite{Shardlow&al'2024}, in which comparable datasets for 10 languages were created. In the guidelines for the data selection it was strongly suggested to use texts from the educational domain.

For Catalan, we could not find a sufficiently large corpus of educational text. So, entences were selected from the Educational news section of the TeCla corpus \cite{armengol-estape-etal-2021-multilingual} of 
news texts. 

For Spanish, we selected educational texts on finance due to their social relevance and the pressing need to make this knowledge accessible to vulnerable populations. Financial literacy, recognized as an essential tool for economic empowerment and inclusion, especially among individuals with disabilities, remains underexplored in text simplification \cite{vieira2023full}. Learning about personal finance is critical in fostering autonomy and improving decision-making. The specialized nature of these texts, characterized by domain-specific terminology and conceptual density, requires careful consideration in simplification approaches to maintain accessibility and accuracy \cite{economico2011educacion}. Our research addresses these challenges by focusing on this area, aligning with broader efforts to promote financial competence and social inclusion for underserved communities.
The Spanish texts originate from publications in South America. 

\end{document}